\documentclass[10pt,twocolumn,letterpaper]{article}

\usepackage{cvpr}
\usepackage{times}
\usepackage{epsfig}
\usepackage{graphicx}
\usepackage{amsmath}
\usepackage{amssymb}
\usepackage{url}
\usepackage{array}
\usepackage{multirow}
\usepackage{tabularx}

\usepackage[british,UKenglish,USenglish,english,american]{babel}

\usepackage[bookmarks=false,colorlinks=true,linkcolor=black,citecolor=black,filecolor=black,urlcolor=black]{hyperref}

\newcommand{\ve}[1]{\mathbf{#1}} 
\newcommand{\ma}[1]{\mathrm{#1}} 
\newcommand{\tr}{{^{\top}}}
\newcommand{\fr}{_\mathrm{F}}
\newcommand{\tabincell}[2]{\begin{tabular}{@{}#1@{}}#2\end{tabular}}

\renewcommand{\arraystretch}{1.01}

\cvprfinalcopy 

\ifcvprfinal\fi
\begin{document}

\title{Efficient and Accurate Approximations of Nonlinear Convolutional Networks}

\author{Xiangyu Zhang$^1$\thanks{This work is done when Xiangyu Zhang and Xiang Ming are interns at Microsoft Research.} \qquad Jianhua  Zou$^1$  \qquad Xiang Ming$^{1*}$ \qquad Kaiming He$^2$ \qquad Jian Sun$^2$\\
$^1$Xi'an Jiaotong University \quad\quad\quad\quad\quad\quad\quad\quad\quad\quad $^2$Microsoft Research
}

\maketitle

\begin{abstract}
This paper aims to accelerate the test-time computation of deep convolutional neural networks (CNNs). Unlike existing methods that are designed for approximating linear filters or linear responses, our method takes the nonlinear units into account. We minimize the reconstruction error of the nonlinear responses, subject to a low-rank constraint which helps to reduce the complexity of filters. We develop an effective solution to this constrained nonlinear optimization problem. An algorithm is also presented for reducing the accumulated error when multiple layers are approximated.
A whole-model speedup ratio of 4$\times$ is demonstrated
on a large network trained for ImageNet, while the top-5 error rate is only increased by 0.9\%. Our accelerated model has a comparably fast speed as the ``AlexNet'' \cite{Krizhevsky2012}, but is 4.7\% more accurate.
\end{abstract}

\section{Introduction}

This paper addresses efficient test-time computation of deep convolutional neural networks (CNNs) \cite{LeCun1989,Krizhevsky2012}. Since the success of CNNs \cite{Krizhevsky2012} for large-scale image classification, the accuracy of the newly developed CNNs \cite{Zeiler2014,Sermanet2014,He2014,Simonyan2014,Szegedy2014} has been continuously improving. However, the computational cost of these networks (especially the more accurate but larger models) also increases significantly. The expensive test-time evaluation of the models can make them impractical in real-world systems. For example, a cloud service needs to process thousands of new requests per seconds; portable devices such as phones and tablets mostly have CPUs or low-end GPUs only; some recognition tasks like object detection \cite{Girshick2014,He2014,Hariharan2014} are still time-consuming for processing a single image even on a high-end GPU. For these reasons and others, it is of practical importance to accelerate the test-time computation of CNNs.

There have been a few studies on approximating deep CNNs for accelerating test-time evaluation \cite{Vanhoucke2011,Denton2014,Jaderberg2014}. A commonly used assumption is that the convolutional filters are approximately low-rank along certain dimensions. So the original filters can be approximately decomposed into a series of smaller filters, and the complexity is reduced. These methods have shown promising speedup ratios on a single \cite{Denton2014} or a few layers \cite{Jaderberg2014} with some degradation of accuracy.

The algorithms and approximations in the previous work are developed for reconstructing linear filters \cite{Denton2014,Jaderberg2014} and linear responses \cite{Jaderberg2014}. However, the nonlinearity like the Rectified Linear Units (ReLU) \cite{Nair2010,Krizhevsky2012} is not involved in their optimization. Ignoring the nonlinearity will impact the quality of the approximated layers. Let us consider a case that the filters are approximated by reconstructing the linear responses. Because the ReLU will follow, the model accuracy is more sensitive to the reconstruction error of the positive responses than to that of the negative responses.

Moreover, it is a challenging task of accelerating the whole network (instead of just one or a very few layers).
The errors will be accumulated if several layers are approximated, especially when the model is deep. Actually, in the recent work \cite{Denton2014,Jaderberg2014} the approximations are applied on a single layer of large CNN models, such as those trained on ImageNet \cite{Deng2009,Russakovsky2014}.
It is insufficient for practical usage to speedup one or a few layers, especially for
the deeper models which have been shown very accurate \cite{Simonyan2014,Szegedy2014,He2014}.

In this paper, a method for accelerating \emph{nonlinear} convolutional networks is proposed.
It is based on minimizing the reconstruction error of \emph{nonlinear} responses, subject to a low-rank constraint that can be used to reduce computation. To solve the challenging constrained optimization problem, we decompose it into two feasible subproblems and iteratively solve them. We further propose to minimize an asymmetric reconstruction error, which effectively reduces the accumulated error of multiple approximated layers.

We evaluate our method on a 7-convolutional-layer model trained on ImageNet. We investigate the cases of accelerating each single layer and the whole model. Experiments show that our method is more accurate than the recent method of Jaderberg \etal's \cite{Jaderberg2014} under the same speedup ratios.
A \emph{whole-model} speedup ratio of 4$\times$ is demonstrated, and its degradation is merely 0.9\%. When our model is accelerated to have a comparably fast speed as the ``AlexNet'' \cite{Krizhevsky2012}, our accuracy is 4.7\% higher.

\section{Approaches}
\label{sec:method}

\subsection{Low-rank Approximation of Responses}
\label{sec:linear}

Our observation is that the response at a position of a convolutional feature map approximately lies on a low-rank subspace. The low-rank decomposition can reduce the complexity. To find the approximate low-rank subspace, we minimize the reconstruction error of the responses.

More formally, we consider a convolutional layer with a filter size of $k\times k\times c$, where $k$ is the spatial size of the filter and $c$ is the number of input channels of this layer. To compute a response, this filter is applied on a $k\times k\times c$ volume of the layer input. We use $\ve{x}\in\mathbb{R}^{k^2c+1}$ to denote a vector that reshapes this volume (appending one as the last entry for the bias). A response $\ve{y}\in\mathbb{R}^{d}$ at a position of a feature map is computed as:
\begin{equation}\label{eq:y}
\ve{y}=\ma{W}\ve{x}.
\end{equation}
where $\ma{W}$ is a $d$-by-($k^2c$$+$$1$) matrix, and $d$ is the number of filters.
Each row of $\ma{W}$ denotes the reshaped form of a $k\times k\times c$ filter (appending the bias as the last entry). We will address the nonlinear case later.

If the vector $\ve{y}$ is on a low-rank subspace, we can write $\ve{y}=\ma{M}(\ve{y}-\bar{\ve{y}})+\bar{\ve{y}}$, where $\ma{M}$ is a $d$-by-$d$ matrix of a rank $d'<d$ and $\bar{\ve{y}}$ is the mean vector of responses. Expanding this equation, we can compute a response by:
\begin{equation}\label{eq:y1}
\ve{y}=\ma{M}\ma{W}\ve{x}+\ve{b},
\end{equation}
where $\ve{b}=\bar{\ve{y}}-\ma{M}\bar{\ve{y}}$ is a new bias. The rank-$d'$ matrix $\ma{M}$ can be decomposed into two $d$-by-$d'$ matrices $\ma{P}$ and $\ma{Q}$ such that $\ma{M}=\ma{P}\ma{Q}\tr$.
We denote $\ma{W}'=\ma{Q}\tr\ma{W}$ as a $d'$-by-($k^2c$$+$$1$) matrix, which is essentially a new set of $d'$ filters. Then we can compute (\ref{eq:y1}) by:
\begin{equation}\label{eq:y2}
\ve{y}=\ma{P}\ma{W}'\ve{x}+\ve{b}.
\end{equation}
The complexity of using Eqn.(\ref{eq:y2}) is $O(d'k^2c)+O(dd')$ , while the complexity of using Eqn.(\ref{eq:y}) is $O(dk^2c)$. For many typical models/layers, we usually have $O(dd')\ll O(d'k^2c)$, so the computation in Eqn.(\ref{eq:y2}) will reduce the complexity to about $d'/d$.

\begin{figure}
\begin{center}
\includegraphics[width=0.8\linewidth]{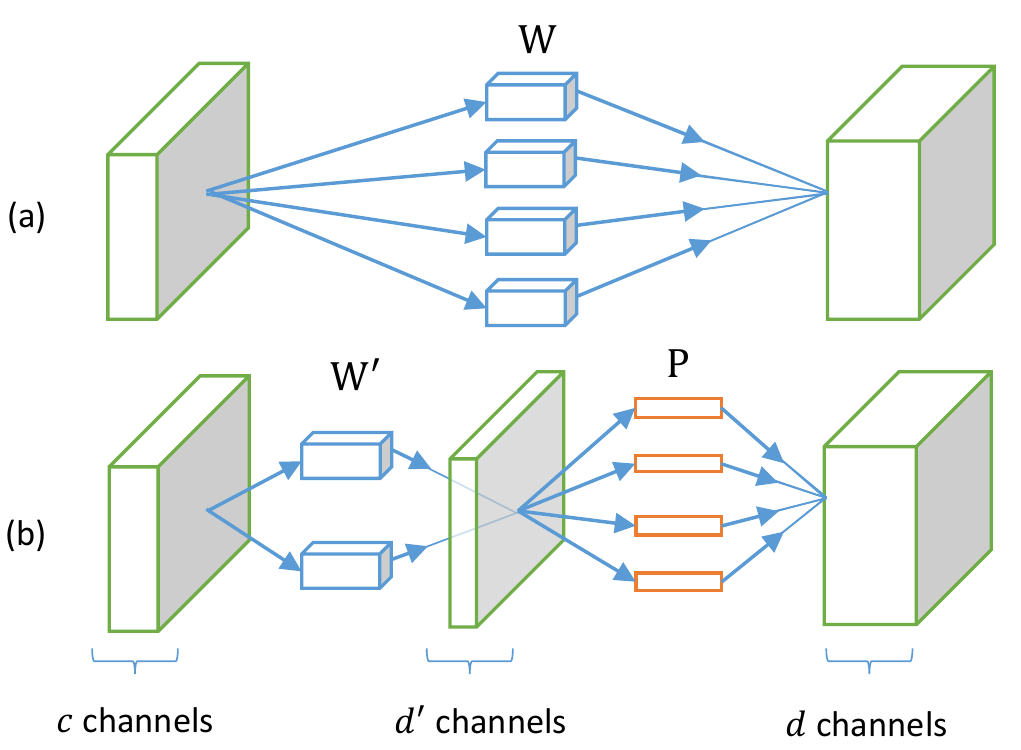}
\end{center}
   \caption{Illustration of the approximation. (a) An original layer with complexity $O(dk^2c)$. (b) An approximated layer with complexity reduced to $O(d'k^2c)+O(dd')$.}
\label{fig:concept}
\end{figure}

Fig.~\ref{fig:concept} illustrates how to use Eqn.(\ref{eq:y2}) in a network. We replace the original layer (given by $\ma{W}$) by two layers (given by $\ma{W}'$ and $\ma{P}$). The matrix $\ma{W}'$ is actually $d'$ filters whose sizes are $k\times k\times c$. These filters produce a $d'$-dimensional feature map. On this feature map, the $d$-by-$d'$ matrix $\ma{P}$ can be implemented as $d$ filters whose sizes are $1\times 1\times d'$. So $\ma{P}$ corresponds to a convolutional layer with a 1$\times$1 spatial support, which maps the $d'$-dimensional feature map to a $d$-dimensional one. The usage of $1\times1$ spatial filters to adjust dimensions has been adopted for designing network architectures \cite{Lin2013,Szegedy2014}. But in those papers, the $1\times1$ filters are used to reduce dimensions, while in our case they restore dimensions.

\begin{figure*}
\begin{center}
\includegraphics[width=0.8\linewidth]{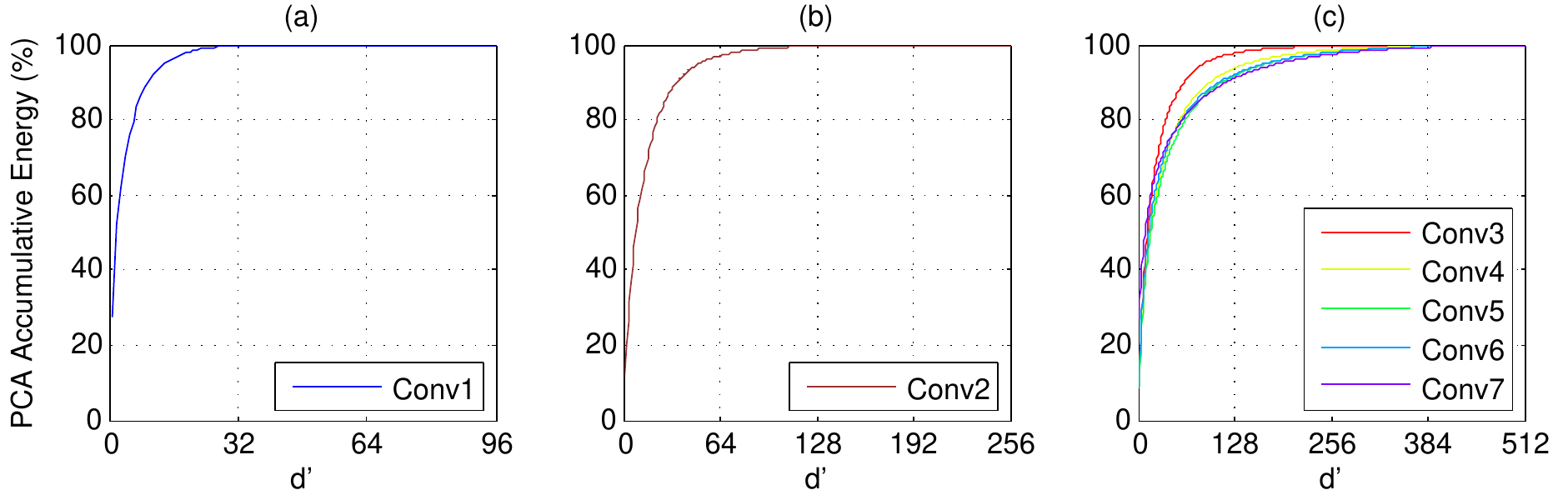}
\end{center}
   \caption{PCA accumulative energy of the responses in each layer, presented as the sum of largest $d'$ eigenvalues (relative to the total energy when $d'=d$). Here the filter number $d$ is 96 for Conv1, 256 for Conv2, and 512 for Conv3-7 (detailed in Table~\ref{tbl:arch}).}
\label{fig:pca_energy}
\end{figure*}

Note that the decomposition of $\ma{M}=\ma{P}\ma{Q}\tr$ can be arbitrary. It does not impact the value of $\ve{y}$ computed in Eqn.(\ref{eq:y2}). A simple decomposition is the Singular Vector Decomposition (SVD) \cite{Golub1996}: $\ma{M}=\ma{U}_{d'}\ma{S}_{d'}\ma{V}_{d'}\tr$, where $\ma{U}_{d'}$ and $\ma{V}_{d'}$ are $d$-by-$d'$ column-orthogonal matrices and $\ma{S}_{d'}$ is a $d'$-by-$d'$ diagonal matrix. Then we can obtain $\ma{P}=\ma{U}_{d'}\ma{S}^{{1}/{2}}_{d'}$ and $\ma{Q}=\ma{V}_{d'}\ma{S}^{{1}/{2}}_{d'}$.

In practice the low-rank assumption is an approximation, and the computation in Eqn.(\ref{eq:y2}) is approximate.
To find an approximate low-rank subspace, we optimize the following problem:
\begin{gather}\label{eq:pca}
\min_{\ma{M}}\sum_i\|(\ve{y}_i-\bar{\ve{y}})-\ma{M}(\ve{y}_i-\bar{\ve{y}})\|^2_2,\\
s.t.\quad rank(\ma{M})\leq d'.\nonumber
\end{gather}
Here $\ve{y}_i$ is a response sampled from the feature maps in the training set.
This problem can be solved by SVD \cite{Golub1996} or actually Principal Component Analysis (PCA): let $\ma{Y}$ be the $d$-by-$n$ matrix concatenating $n$ responses with the mean subtracted, compute the eigen-decomposition of the covariance matrix $\ma{Y}\ma{Y}\tr=\ma{U}\ma{S}\ma{U}\tr$ where $\ma{U}$ is an orthogonal matrix and $\ma{S}$ is diagonal, and $\ma{M}=\ma{U}_{d'}\ma{U}_{d'}\tr$ where $\ma{U}_{d'}$ are the first $d'$ eigenvectors. With the matrix $\ma{M}$ computed, we can find $\ma{P}=\ma{Q}=\ma{U}_{d'}$.

How good is the low-rank assumption of the responses? We sample the responses from a CNN model (with 7 convolutional layers, detailed in Sec.~\ref{sec:exp}) trained on ImageNet \cite{Deng2009}. For the responses of a convolutional layer (from 3,000 randomly sampled training images), we compute the eigenvalues of their covariance matrix and then plot the sum of the largest eigenvalues (Fig.~\ref{fig:pca_energy}). We see that substantial energy is in a small portion of the largest eigenvectors. For example, in the Conv2 layer ($d=256$) the first 128 eigenvectors contribute over 99.9\% energy; in the Conv7 layer ($d=512$), the first 256 eigenvectors contribute over 95\% energy.
This indicates that we can use a fraction of the filters to precisely approximate the original filters.

The low-rank behavior of the responses $\ve{y}$ is because of the low-rank behaviors of the filters $\ma{W}$ and the inputs $\ve{x}$. While the low-rank assumptions of filters have been adopted in recent work \cite{Denton2014,Jaderberg2014}, we further adopt the low-rank assumptions of the filter input $\ve{x}$, which is a local volume and should have correlations. The responses $\ve{y}$ will have lower rank than $\ma{W}$ and $\ma{x}$, so the approximation can be more precise.
In our optimization (\ref{eq:pca}), we directly address the low-rank subspace of $\ve{y}$.

\subsection{The Nonlinear Case}
\label{sec:nonlinear}

Next we investigate the case of using nonlinear units. We use $r(\cdot)$ to denote the nonlinear operator. In this paper we focus on the Rectified Linear Unit (ReLU) \cite{Nair2010}: $r(\cdot)=\max(\cdot, 0)$.
A nonlinear response is given by $r(\ma{W}\ve{x})$ or simply $r(\ve{y})$. We minimize the reconstruction error of the nonlinear responses:
\begin{gather}\label{eq:relu}
\min_{\ma{M},\ve{b}}\sum_i\|r(\ve{y}_i)-r(\ma{M}\ve{y}_i+\ve{b})\|^2_2,\\
s.t.\quad rank(\ma{M})\leq d'.\nonumber
\end{gather}
Here $\ve{b}$ is a new bias to be optimized, and $r(\ma{M}\ve{y}+\ve{b})=r(\ma{M}\ma{W}\ve{x}+\ve{b})$ is the nonlinear response computed by the approximated filters.

The above problem is challenging due to the nonlinearity and the low-rank constraint.
To find a feasible solution, we relax it as:
\begin{gather}
\min_{\ma{M},\ve{b},\{\ve{z}_i\}}\sum_i\|r(\ve{y}_i)-r(\ve{z}_i)\|^2_2
+\lambda\|\ve{z}_i-(\ma{M}\ve{y}_i+\ve{b})\|^2_2\nonumber\\
s.t.\quad rank(\ma{M})\leq d'.
\label{eq:relu1}
\end{gather}
Here $\{\ve{z}_i\}$ is a set of auxiliary variables of the same size as $\{\ve{y}_i\}$. $\lambda$ is a penalty parameter. If $\lambda\rightarrow \infty$, the solution to (\ref{eq:relu1}) will converge to the solution to (\ref{eq:relu}) \cite{Wang2010}. We adopt an alternating solver, fixing $\{\ve{z}_i\}$ and solving for $\ma{M}$, $\ve{b}$ and vice versa.

\vspace{8pt}
\noindent \textbf{(i) The subproblem of $\ma{M}$, $\ve{b}$}. In this case, $\{\ve{z}_i\}$ are fixed.
It is easy to show $\ve{b}=\bar{\ve{z}}-\ma{M}\bar{\ve{y}}$ where $\bar{\ve{z}}$ is the sample mean of $\{\ve{z}_i\}$. Substituting $\ve{b}$ into the objective function, we obtain the problem involving $\ma{M}$:
\begin{gather}\label{eq:pcaz}
\min_{\ma{M}}\sum_i\|(\ve{z}_i-\bar{\ve{z}})-\ma{M}(\ve{y}_i-\bar{\ve{y}})\|^2_2,\\
s.t.\quad rank(\ma{M})\leq d'.\nonumber
\end{gather}
Let $\ma{Z}$ be the $d$-by-$n$ matrix concatenating the vectors of $\{\ve{z}_i-\bar{\ve{z}}\}$. We rewrite the above problem as:
\begin{gather}\label{eq:rank}
\min_{\ma{M}}\|\ma{Z}-\ma{M}\ma{Y}\|^2\fr,\\
s.t.\quad rank(\ma{M})\leq d'.\nonumber
\end{gather}
Here $\|\cdot\|\fr$ is the Frobenius norm.
This optimization problem is a Reduced Rank Regression problem \cite{Gower2004,Takane2006,Takane2007}, and it can be solved by a kind of Generalized Singular Vector Decomposition (GSVD) \cite{Gower2004,Takane2006,Takane2007}. The solution is as follows. Let $\ma{\hat{M}}=\ma{Z}\ma{Y}\tr(\ma{Y}\ma{Y}\tr)^{-1}$. The GSVD is applied on $\ma{\hat{M}}$ as $\ma{\hat{M}}=\ma{U}\ma{S}\ma{V}\tr$, such that $\ma{U}$ is a $d$-by-$d$ orthogonal matrix satisfying $\ma{U}\tr\ma{U}=\ma{I}_d$ where $\ma{I}_d$ is a $d$-by-$d$ identity matrix, and
$\ma{V}$ is a $d$-by-$d$ matrix satisfying $\ma{V}\tr \ma{Y}\ma{Y}\tr \ma{V=\ma{I}_d}$ (called \emph{generalized orthogonality}).
Then the solution $\ma{M}$ to (\ref{eq:rank}) is given by $\ma{M}=\ma{U}_{d'}\ma{S}_{d'}\ma{V}_{d'}\tr$ where $\ma{U}_{d'}$ and $\ma{V}_{d'}$ are the first $d'$ columns of $\ma{U}$ and $\ma{V}$ and $\ma{S}_{d'}$ are the largest $d'$ singular values. We can further show that if $\ma{Z}=\ma{Y}$ (so the problem in (\ref{eq:pcaz}) becomes (\ref{eq:pca})), this solution degrades to computing the eigen-decomposition of $\ma{Y}\ma{Y}\tr$.

\vspace{8pt}
\noindent \textbf{(ii) The subproblem of $\{\ve{z}_i\}$}.
In this case, $\ma{M}$ and $\ve{b}$ are fixed. Then in this subproblem each element $z_{ij}$ of each vector $\ve{z}_i$ is independent of any other. So we solve a 1-dimensional optimization problem as follows:
\begin{gather}
\min_{z_{ij}}~(r(y_{ij})-r(z_{ij}))^2+\lambda(z_{ij}-y'_{ij})^2,
\label{eq:relu2}
\end{gather}
where $y'_{ij}$ is the $j$-th entry of $\ma{M}\ve{y}_i+\ve{b}$. We can separately consider $z_{ij}\geq0$ and $z_{ij}<0$ and remove the ReLU operator. Then we can derive the solution as follows: let
\begin{gather}
z_{ij}^{'} = \min (0, y'_{ij})\\
z_{ij}^{''} = \max (0, \frac{\lambda\cdot y'_{ij} + r(y_{ij})}{\lambda + 1})
\end{gather}
then $z_{ij}=z_{ij}^{'}$ if $z_{ij}^{'}$ gives a smaller value in (\ref{eq:relu2}) than $z_{ij}^{''}$, and otherwise $z_{ij}=z_{ij}^{''}$.

Although we focus on the ReLU, our method is applicable for other types of nonlinearities. The subproblem in (\ref{eq:relu2}) is a 1-dimensional nonlinear least squares problem, so can be solved by gradient descent or simply line search. We plan to study this issue in the future.

\vspace{12pt}
We alternatively solve (i) and (ii).
The initialization is given by the solution to the linear case (\ref{eq:pca}). We warm up the solver by setting the penalty parameter $\lambda=0.01$ and run 25 iterations. Then we increase the value of $\lambda$. In theory,
$\lambda$ should be gradually increased to infinity \cite{Wang2010}. But
we find that it is difficult for the iterative solver to make progress if $\lambda$ is too large. So we increase $\lambda$ to 1, run 25 more iterations, and use the resulting $\ma{M}$ as our solution.
Then we compute $\ma{P}$ and $\ma{Q}$ by SVD on $\ma{M}$.

\subsection{Asymmetric Reconstruction for Multi-Layer}
\label{sec:asymmetric}

To accelerate a whole network, we apply the above method sequentially on each layer, from the shallow layers to the deeper ones. If a previous layer is approximated, its error can be accumulated when the next layer is approximated. We propose an asymmetric reconstruction method to address this issue.

Let us consider a layer whose input feature map is not precise due to the approximation of the previous layer/layers. We denote the approximate input to the current layer as $\ve{\hat{x}}$. For the training samples, we can still compute its non-approximate responses as $\ve{y}=\ma{W}\ve{x}$. So we can optimize an ``asymmetric'' version of (\ref{eq:relu}):
\begin{gather}\label{eq:reluasy}
\min_{\ma{M},\ve{b}}\sum_i\|r(\ma{W}\ve{x}_i)-r(\ma{M}\ma{W}\ve{\hat{x}}_i+\ve{b})\|^2_2,\\
s.t.\quad rank(\ma{M})\leq d'.\nonumber
\end{gather}
Here in the first term $\ve{x}_i$ is the non-approximate input, while in the second term $\ve{\hat{x}}_i$ is the approximate input due to the previous layer. We need not use $\ve{\hat{x}}_i$ in the first term, because $r(\ma{W}\ve{x}_i)$ is the real outcome of the original network and thus is more precise. On the other hand, we do not use $\ve{x}_i$ in the second term, because $r(\ma{M}\ma{W}\ve{\hat{x}}_i+\ve{b})$ is the actual operation of the approximated layer. This asymmetric version can reduce the accumulative errors when multiple layers are approximated. The optimization problem in (\ref{eq:reluasy}) can be solved using the same algorithm as for (\ref{eq:relu}).

\begin{table*}
\begin{center}
\small
\begin{tabular}{|c|c|c|c|c|c|c|c|c|}
\hline
layer & filter size & \# channels & \# filters & stride & output size & complexity (\%) & \# of zeros \\
\hline
Conv1 & 7 $\times$ 7 & 3 & 96 & 2 & 109 $\times$ 109 & 3.8 & 0.49 \\
Pool1 & 3 $\times$ 3 &   &    & 3 & 37 $\times$ 37   &     &      \\
\hline
Conv2 & 5 $\times$ 5 & 96 & 256 & 1 & 35 $\times$ 35 & 17.3 & 0.62 \\
Pool2 & 2 $\times$ 2 &   &    & 2 & 18 $\times$ 18   &     &      \\
\hline
Conv3 & 3 $\times$ 3 & 256 & 512 & 1 & 18 $\times$ 18 & 8.8 & 0.60 \\
Conv4 & 3 $\times$ 3 & 512 & 512 & 1 & 18 $\times$ 18 & 17.5 & 0.69 \\
Conv5 & 3 $\times$ 3 & 512 & 512 & 1 & 18 $\times$ 18 & 17.5 & 0.69 \\
Conv6 & 3 $\times$ 3 & 512 & 512 & 1 & 18 $\times$ 18 & 17.5 & 0.68 \\
Conv7 & 3 $\times$ 3 & 512 & 512 & 1 & 18 $\times$ 18 & 17.5 & 0.95 \\
\hline
\end{tabular}
\end{center}
\caption{The architecture of the model. Each convolutional layer is followed by ReLU. The final convolutional layer is followed by a spatial pyramid pooling layer \cite{He2014} that have 4 levels ($\{6\times6, 3\times3, 2\times2, 1\times1\}$, totally 50 bins). The resulting $50\times512$-d is fed into the 4096-d fc layer (fc6), followed by another 4096-d fc layer (fc7) and a 1000-way softmax layer. The convolutional complexity is the theoretical time complexity, shown as relative numbers to the total convolutional complexity. The (relative) number of zeros is the calculated on the responses of the layer, which shows the ``sparsity'' of the layer.}
\label{tbl:arch}
\end{table*}

\begin{figure}
\begin{center}
\includegraphics[width=0.85\linewidth]{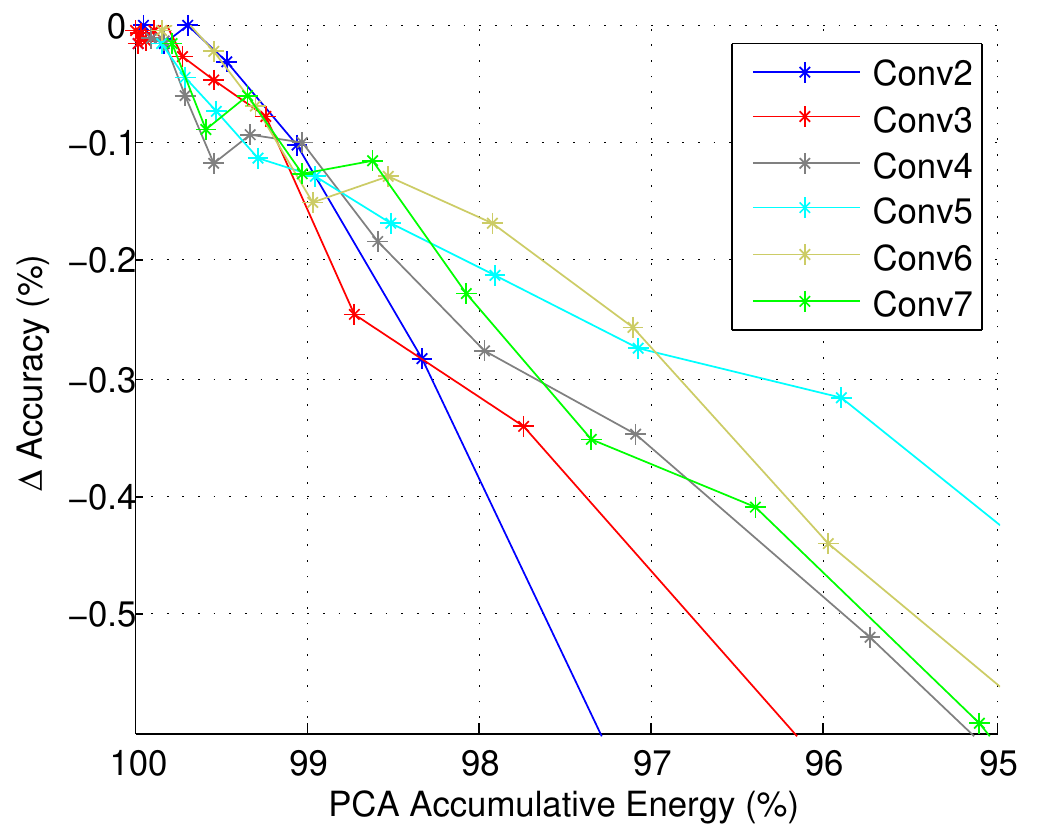}
\end{center}
   \caption{PCA accumulative energy and the accuracy rates (top-5). Here the accuracy is evaluated using the linear solution (the nonlinear solution has a similar trend). Each layer is evaluated independently, with other layers not approximated. The accuracy is shown as the difference to no approximation.}
\label{fig:pca_sigma}
\end{figure}

\subsection{Rank Selection for Whole-Model Acceleration}
\label{sec:parameterselection}

In the above, the optimization is based on a target $d'$ of each layer. $d'$ is the only parameter that determines the complexity of an accelerated layer.
But given a desired speedup ratio of the \emph{whole model}, we need to determine the proper rank $d'$ used for each layer.

Our strategy is based on an empirical observation that the PCA energy is related to the classification accuracy after approximations. To verify this observation, in Fig.~\ref{fig:pca_sigma} we show the classification accuracy (represented as the difference to no approximation) \vs the PCA energy.
Each point in this figure is empirically evaluated using a value of $d'$. 100\% energy means no approximation and thus no degradation of classification accuracy.
Fig.~\ref{fig:pca_sigma} shows that the classification accuracy is roughly linear on the PCA energy.

To simultaneously determine the rank for each layer, we further assume that the whole-model classification accuracy is roughly related to the product of the PCA energy of all layers. More formally, we consider this objective function:
\begin{gather}
\mathcal{E}=\prod_{l}\sum_{a=1}^{d'_{l}}{\sigma_{l,a}}
\end{gather}
Here $\sigma_{l,a}$ is the $a$-th largest eigenvalue of the layer $l$, and $\sum_{a=1}^{d'_{l}}{\sigma_{l,a}}$ is the PCA energy of the largest $d'_{l}$ eigenvalues in the layer $l$. The product $\prod_{l}$ is over all layers to be approximated. The objective $\mathcal{E}$ is assumed to be related to the accuracy of the approximated whole network.
Then we optimize this problem:
\begin{gather}\label{eq:compcost}
\max_{\{d'_l\}}\mathcal{E},\quad\quad
s.t.\quad \sum_l{\frac{d'_l}{d_l}C_l} \leq C.
\end{gather}
Here $d_l$ is the original number of filters in the layer $l$, and $C_l$ is the original time complexity of the layer $l$. So $\frac{d'_l}{d_l}C_l$ is the complexity after the approximation. $C$ is the total complexity after the approximation, which is given by the desired speedup ratio. This problem means that we want to maximize the accumulated accuracy subject to the time complexity constraint.

The problem in (\ref{eq:compcost}) is a combinatorial problem \cite{Reeves1993}.
So we adopt a greedy strategy to solve it. We initialize $d'_l$ as $d_l$, and consider the set $\{\sigma_{l,a}\}$. In each step we remove an eigenvalue $\sigma_{l,d'_l}$ from this set, chosen from a certain layer $l$. The relative reduction of the objective is $\triangle \mathcal{E} / \mathcal{E}=\sigma_{l,d'}/{\sum_{a=1}^{d'_l}\sigma_{l,a}}$, and the reduction of complexity is $\triangle C = {\frac{1}{d_l}C_l}$. Then we define a measure as $\frac{\triangle \mathcal{E} / \mathcal{E}}{\triangle C}$.
The eigenvalue $\sigma_{l,d'_l}$ that has the smallest value of this measure is removed. Intuitively, this measure favors a small reduction of $\triangle \mathcal{E} / \mathcal{E}$ and a large reduction of complexity $\triangle C$. This step is greedily iterated, until the constraint of the total complexity is achieved.

\begin{figure*}[t]
\begin{center}
\includegraphics[width=0.85\linewidth]{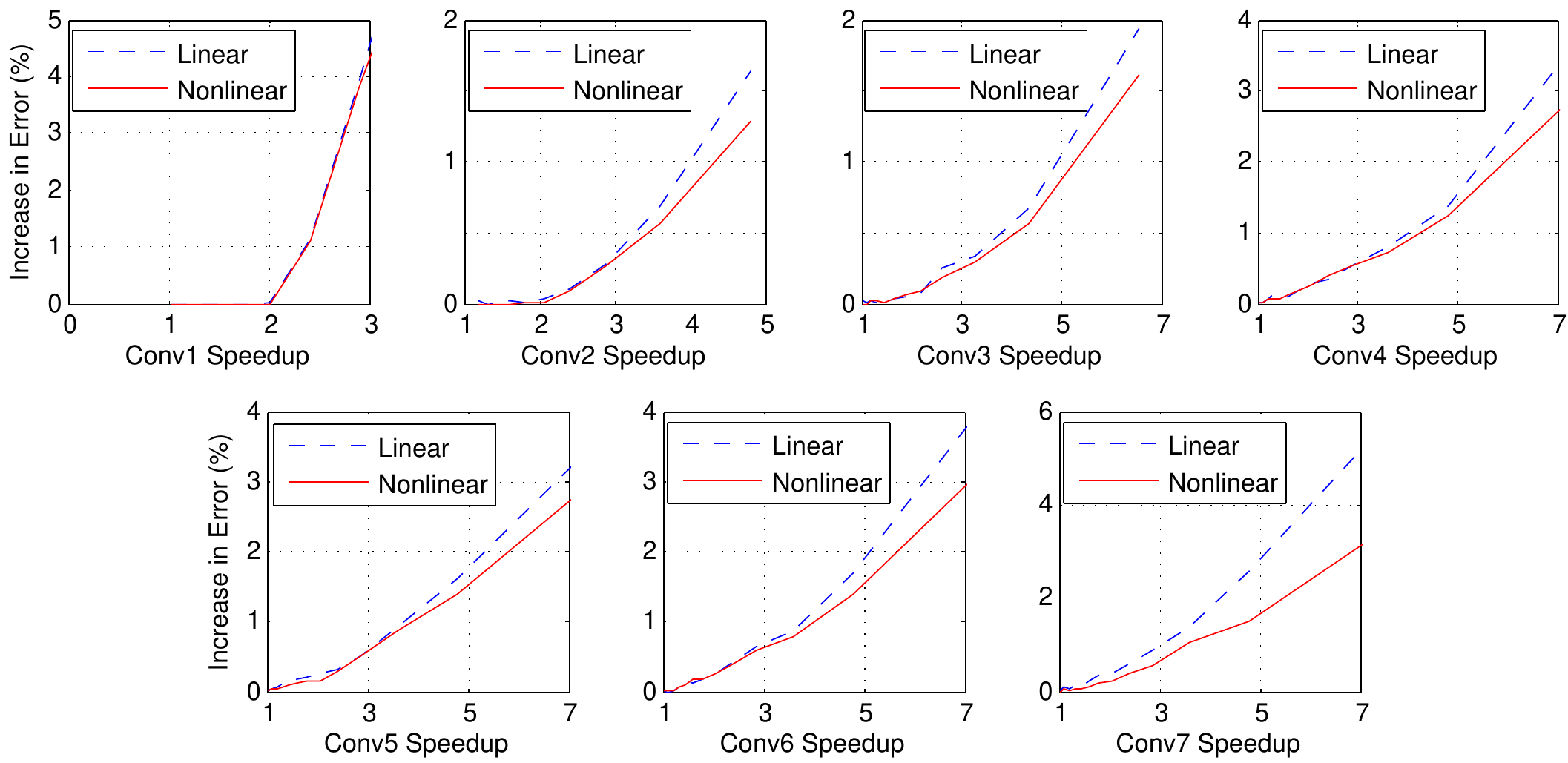}
\end{center}
   \caption{\textbf{Linear vs. Nonlinear}: single-layer performance of accelerating Conv1 to Conv7. The speedup ratios are computed by the theoretical complexity, but is nearly the same as the actual speedup ratios in our CPU/GPU implementation. The error rates are top-5 single-view, and shown as the increase of error rates compared with no approximation (\emph{smaller is better}).}
\label{fig:layerwiseresult}
\end{figure*}

\begin{figure*}[t]
\begin{center}
\includegraphics[width=0.8\linewidth]{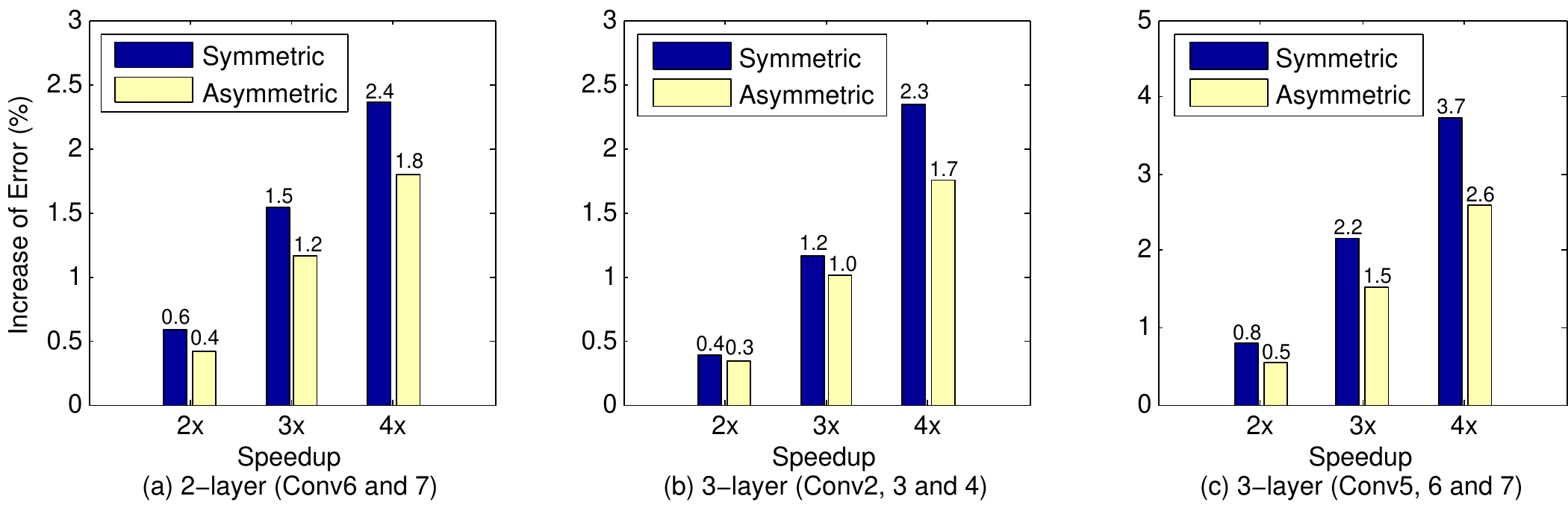}
\end{center}
   \caption{\textbf{Symmetric vs. Asymmetric}: the cases of 2-layer and 3-layer approximation. The speedup is computed by the complexity of the layers approximated. (a) Approximation of Conv6 \& 7. (b) Approximation of Conv2, 3 \& 4. (c) Approximation of Conv5, 6 \& 7.}
\label{fig:asym}
\end{figure*}

\subsection{Discussion}
\label{sec:discussion}

In our formulation, we focus on reducing the number of filters (from $d$ to $d'$).
There are algorithmic advantages of operating on the ``$d$'' dimension. Firstly, this dimension can be easily controlled by the rank constraint $rank(\ma{M})\leq d'$. This constraint enables closed-form solutions, \eg, PCA to the problem (\ref{eq:pca}) or GSVD to the subproblem (\ref{eq:pcaz}).
Secondly, the optimized low-rank projection $\ma{M}$ can be exactly decomposed into low-dimensional filters ($\ma{P}$ and $\ma{Q}$) by SVD. These simple and close-form solutions can produce good results using a very small subset of training images (3,000 out of one million).

\section{Experiments}
\label{sec:exp}

We evaluate on the ``SPPnet (Overfeat-7)'' model \cite{He2014}, which is one of the state-of-the-art models for ImageNet Large Scale Visual Recognition Challenge (ILSVRC) 2014 \cite{Russakovsky2014}.
This model (detailed in Table~\ref{tbl:arch}) has a similar architecture to the Overfeat model \cite{Sermanet2014}, but has 7 convolutional layers.
A spatial pyramid pooling layer \cite{He2014} is used after the last convolutional layer, which improves the classification accuracy. We train the model on the 1000-class dataset of ImageNet 2012 \cite{Deng2009,Russakovsky2014}, following the details in \cite{He2014}.

We evaluate the ``top-5 error'' (or simply termed as ``error'') using single-view testing. The view is the center $224\times224$ region cropped from the resized image whose shorter side is 256. The single-view error rate of the model is 12.51\% on the ImageNet validation set, and the increased error rates of the approximated models are all based on this number. For completeness, we report that this model has 11.1\% error using 10-view test and 9.3\% using 98-view test.

We use this model due to the following reasons. First, its architecture is similar to many existing models \cite{Krizhevsky2012,Zeiler2014,Sermanet2014,Chatfield2014} (such as the first/second layers and the cascade usage of $3\times3$ filters), so we believe most observations should be valid on other models. Second, on the other hand, this model is deep (7-conv.) and the computation is more uniformly distributed among the layers (see ``complexity'' in Table~\ref{tbl:arch}). A similar behavior exhibits on the compelling VGG-16/19 models \cite{Simonyan2014}. The uniformly distributed computation indicates that most layers should be accelerated for an overall speedup.

For the training of the approximations as in (\ref{eq:pca}), (\ref{eq:relu1}), and (\ref{eq:reluasy}), we randomly sample 3,000 images from the ImageNet training set and use their responses as the training samples.

\subsection{Single-Layer: Linear \vs Nonlinear}
\label{sec:layerwise}

In this subsection we evaluate the single-layer performance. When evaluating a single approximated layer, the rest layers are unchanged and not approximated. The speedup ratio (involving that single layer only) is shown as the theoretical ratio computed by the complexity.

In Fig.~\ref{fig:layerwiseresult} we compare the performance of our linear solution (\ref{eq:pca}) and nonlinear solution (\ref{eq:relu1}). The performance is displayed as \emph{increase of error rates} (decrease of accuracy) \vs the speedup ratio of that layer. Fig.~\ref{fig:layerwiseresult} shows that the nonlinear solution consistently performs better than the linear solution. In Table~\ref{tbl:arch}, we show the sparsity (the portion of zero activations after ReLU) of each layer. A zero activation is due to the truncation of ReLU. The sparsity is over 60\% for Conv2-7, indicating that the ReLU takes effect on a substantial portion of activations. This explains the discrepancy between the linear and nonlinear solutions. Especially, the Conv7 layer has a sparsity of 95\%, so the advantage of the nonlinear solution is more obvious.

Fig.~\ref{fig:layerwiseresult} also shows that when accelerating only a single layer by 2$\times$, the increased error rates of our solutions are rather marginal or ignorable. For the Conv2 layer, the error rate is increased by $<0.1\%$; for the Conv3-7 layers, the error rate is increased by $<0.2\%$.

We also notice that for Conv1, the degradation is ignorable on or below $2\times$ speedup ($1.8\times$ corresponds to $d'=32$). This can be explained by Fig.~\ref{fig:pca_energy}(a): the PCA energy has almost no loss when $d'\geq32$.
But the degradation can grow quickly for larger speedup ratios, because in this layer the channel number $c=3$ is small and $d'$ needs to be reduced drastically to achieve the speedup ratio. So in the following, we will use $d'=32$ for Conv1.

\begin{table*}[t]
\begin{center}
\small
\begin{tabular}{|c|c|ccccccc|c|}
\hline
speedup & rank sel. & Conv1 & Conv2 & Conv3 & Conv4 & Conv5 & Conv6 & Conv7 & err. $\uparrow \%$\\
\hline\hline
2$\times$ & no & 32 & 110 & 199 & 219 & 219 & 219 & 219 & 1.18 \\
2$\times$ & \textbf{yes} & 32 & 83 & 182 & 211 & 239 & 237 & 253 & \textbf{0.93} \\
\hline
2.4$\times$ & no & 32 & 96 & 174 & 191 & 191 & 191 & 191 & 1.77 \\
2.4$\times$ & \textbf{yes} & 32 & 74 & 162 & 187 & 207 & 205 & 219 & \textbf{1.35} \\
\hline
3$\times$ & no & 32 & 77 & 139 & 153 & 153 & 153 & 153 & 2.56 \\
3$\times$ & \textbf{yes} & 32 & 62 & 138 & 149 & 166 & 162 & 167 & \textbf{2.34} \\
\hline
4$\times$ & no & 32 & 57 & 104 & 115 & 115 & 115 & 115 & 4.32 \\
4$\times$ & \textbf{yes} & 32 & 50 & 112 & 114 & 122 & 117 & 119 & \textbf{4.20} \\
\hline
5$\times$ & no & 32 & 46 & 83 & 92 & 92 & 92 & 92 & 6.53 \\
5$\times$ & \textbf{yes} & 32 & 41 & 94 & 93 & 98 & 92 & 90 & \textbf{6.47} \\
\hline
\end{tabular}
\end{center}
\caption{\textbf{Whole-model acceleration with/without rank selection}. The speedup ratios shown here involve all convolutional layers (Conv1-Conv7). We fix $d'=32$ in Conv1.
In the case of no rank selection, the speedup ratio of each other layer is the same. The solver is the asymmetric version. Each column of Conv1-7 shows the rank $d'$ used, which is the number of filters after approximation. The error rates are top-5 single-view, and shown as the increase of error rates compared with no approximation (\emph{smaller is better}).}
\label{tbl:entire_speedup}
\end{table*}

\subsection{Multi-Layer: Symmetric \vs Asymmetric}

Next we evaluate the performance of asymmetric reconstruction as in the problem (\ref{eq:reluasy}). We demonstrate approximating 2 layers or 3 layers. In the case of 2 layers, we show the results of approximating Conv6 and 7; and in the case of 3 layers, we show the results of approximating Conv5-7 or Conv2-4. The comparisons are consistently observed for other cases of multi-layer.

We sequentially approximate the layers involved, from a shallower one to a deeper one. In the asymmetric version (\ref{eq:reluasy}), $\ve{\hat{x}}$ is from the output of the previous approximated layer (if any), and $\ve{x}$ is from the output of the previous non-approximate layer. In the symmetric version (\ref{eq:relu}), the response $\ve{y}=\ma{M}\ve{x}$ where $\ve{x}$ is from the output of the previous non-approximate layer.
We have also tried another symmetric version of $\ve{y}=\ma{M}\ve{\hat{x}}$ where $\ve{\hat{x}}$ is from the output of the previous approximated layer (if any), and found this symmetric version is even worse.

Fig.~\ref{fig:asym} shows the comparisons between the symmetric and asymmetric versions. The asymmetric solution has significant improvement over the symmetric solution. For example, when only 3 layers are approximated simultaneously (like Fig.~\ref{fig:asym} (c)), the improvement is over 1.0\% when the speedup is 4$\times$.
This indicates that the accumulative error rate due to multi-layer approximation can be effectively reduced by the asymmetric version.

When more and all layers are approximated simultaneously (as below), if without the asymmetric solution, the error rates will increase more drastically.

\subsection{Whole-Model: with/without Rank Selection}

In Table~\ref{tbl:entire_speedup} we show the results of whole-model acceleration. The solver is the asymmetric version.
For Conv1, we fix $d'=32$. For other layers,
when the rank selection is not used, we adopt the same speedup ratio on each layer and determine its desired rank $d'$ accordingly. When the rank selection is used, we apply it to select $d'$ for Conv2-7.
Table~\ref{tbl:entire_speedup} shows that the rank selection consistently outperforms the counterpart without rank selection. The advantage of rank selection is observed in both linear and nonlinear solutions.

In Table~\ref{tbl:entire_speedup} we notice that the rank selection often chooses a higher rank $d'$ (than the no rank selection) in Conv5-7. For example, when the speedup is 3$\times$, the rank selection assigns $d'=167$ to Conv7, while this layer only requires $d'=153$ to achieve 3$\times$ single-layer speedup of itself. This can be explained by Fig.~\ref{fig:pca_energy}(c). The energy of Conv5-7 is less concentrated, so these layers require higher ranks to achieve good approximations.

\subsection{Comparisons with Previous Work}

We compare with Jaderberg \etal's method \cite{Jaderberg2014}, which is a recent state-of-the-art solution to efficient evaluation. This method mainly operates on the spatial domain. It decomposes a $k\times k$ spatial support into a cascade of $k\times 1$ and $1\times k$ spatial supports. This method focuses on the linear reconstruction error. The SGD solver is adopted for optimization. In the paper of \cite{Jaderberg2014}, their method is only evaluated on a single layer of a model trained for ImageNet.

Our comparisons are based on our re-implementation of \cite{Jaderberg2014}. We use the \emph{Scheme 2} decomposition in \cite{Jaderberg2014} and its filter reconstruction version, which is the one used for ImageNet as in \cite{Jaderberg2014}.
Our re-implementation of \cite{Jaderberg2014} gives a 2$\times$ single-layer speedup on Conv2 and $<0.2\%$ increase of error. As a comparison, in \cite{Jaderberg2014} it reports $0.5\%$ increase of error on Conv2 under a 2$\times$ single-layer speedup, evaluated on another Overfeat model \cite{Sermanet2014}.
For whole-model speedup, we adopt this method sequentially on Conv2-7 using the same speedup ratio. We do not apply this method on Conv1, because this layer has a small fraction of complexity while the spatial decomposition leads to considerable error on this layer if using a speedup ratio similar to other layers.

\begin{figure}[t]
\begin{center}
\includegraphics[width=0.8\linewidth]{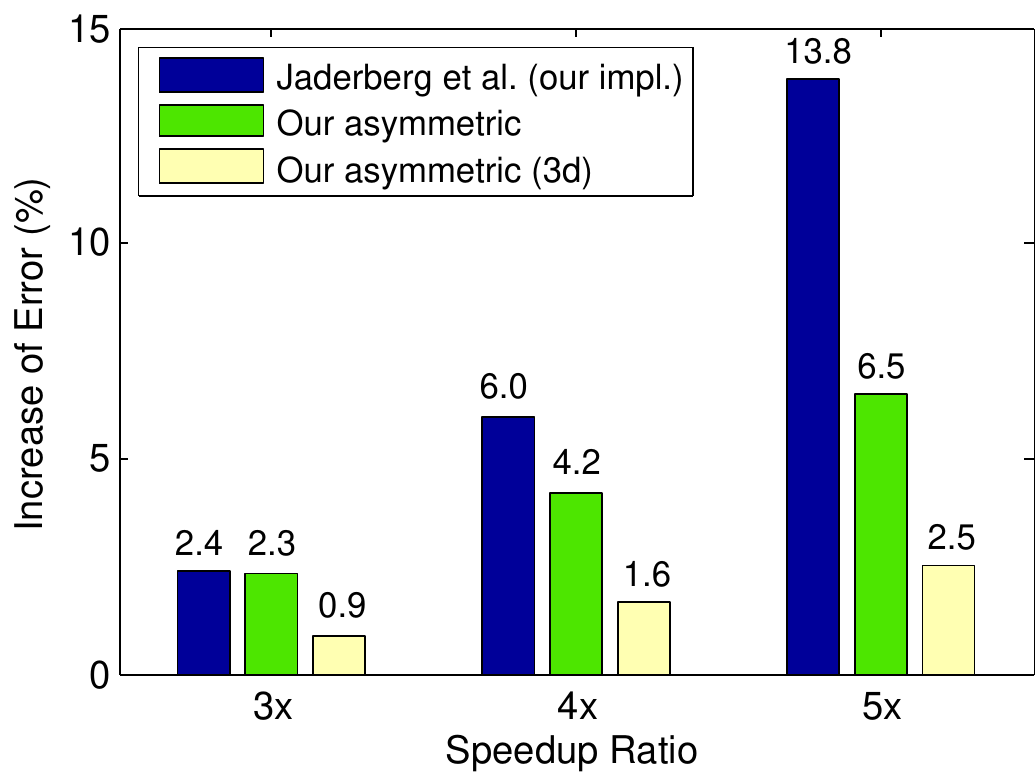}
\end{center}
   \caption{Comparisons with Jaderberg \etal's spatial decomposition method \cite{Jaderberg2014}. The error rates are top-5 single-view, and shown as \textbf{the increase of error rates} compared with no approximation (\emph{smaller is better}).}
\label{fig:combine_vgg_entire}
\end{figure}

In Fig.~\ref{fig:combine_vgg_entire} we compare our method with Jaderberg \etal's \cite{Jaderberg2014} for whole-model speedup. The speedup ratios are the theoretical complexity ratios involving all convolutional layers. Our method is the asymmetric version and with rank selection (denoted as ``\emph{our asymmetric}''). Fig.~\ref{fig:combine_vgg_entire} shows that when the speedup ratios are large (4$\times$ and 5$\times$), our method outperforms Jaderberg \etal's method significantly. For example, when the speedup ratio is 4$\times$, the increased error rate of our method is 4.2\%, while Jaderberg \etal's is 6.0\%. Jaderberg \etal's result degrades quickly when the speedup ratio is getting large, while ours degrades more slowly. This is indicates the effects of our method for reducing accumulative error.
In our CPU implementation, both methods have similar actual speedup ratios for a given theoretical speedup, for example, 3.55$\times$ actual for 4$\times$ theoretical speedup. It is because the overhead for both methods mainly comes from the fully-connected and other layers.

Because our asymmetric solution can effectively reduce the accumulated error, we can approximate a layer by the two methods simultaneously, and the asymmetric reconstruction of the next layer can reduce the error accumulated by the two methods.
As discussed in Sec.~\ref{sec:discussion}, our method is based on the channel dimension ($d$), while Jaderberg \etal's method mainly exploits the decomposition of the two spatial dimensions. These two mechanisms are complementary, so
we conduct the following sequential strategy.
The Conv1 layer is approximated using our model only. Then for the Conv2 layer, we first apply our method. The approximated layer has $d'$ filters whose sizes are $k \times k \times c$ followed by $1\times1$ filters (as in Fig.~\ref{fig:concept}(b)). Next we apply Jaderberg \etal's method to decompose the spatial support into a cascade of $k\times1$ and $1\times k$ filters (\emph{Scheme 2} \cite{Jaderberg2014}). This gives a 3-dimensional approximation of Conv2. Then we apply our method on Conv3.
Now the asymmetric solver will take the responses approximated by the two mechanisms as the input, while the reconstruction target is still the responses of the original network. So while Conv2 has been approximated twice, the asymmetric solver of Conv3 can partially reduce the accumulated error. This process is sequentially adopted in the layers that follow.

In Fig.~\ref{fig:combine_vgg_entire} we show the results of this 3-dimensional decomposition strategy (denoted as ``\emph{our asymmetric (3d)}''). We set the speedup ratios of both mechanisms to be equal: \eg, if the speedup ratio of the whole model is $r\times$, then we use $\sqrt r\times$ for both.
Fig.~\ref{fig:combine_vgg_entire} shows that
this strategy leads to significantly smaller increase of error. For example, when the speedup is 5$\times$, the error is increased by only 2.5\%. This is because the speedup ratio is accounted by all three dimensions, and the reduction of each dimension is lower. Our asymmetric solver effectively controls the accumulative error even if the multiple layers are decomposed extensively.

Finally, we compare the accelerated whole model with the well-known ``\emph{AlexNet}'' \cite{Krizhevsky2012}. The comparison is based on our re-implementation of AlexNet. The architecture is the same as in \cite{Krizhevsky2012} except that the GPU splitting is ignored. Besides the standard strategies used in \cite{Krizhevsky2012}, we train this model using the 224$\times$224 views cropped from resized images whose shorter edge is 256 \cite{Howard2013}. Our re-implementation of this model has top-5 single-view error rate as 18.8\% (10-view top-5 16.0\% and top-1 37.6\%). This is better than the one reported in \cite{Krizhevsky2012}\footnote{In \cite{Krizhevsky2012} the 10-view error is top-5 18.2\% and top-1 40.7\%.}.

Table~\ref{tbl:alexnet} shows the comparisons on the accelerated models and AlexNet. The error rates in this table are the absolute value (not the increased number). The time is the actual running time per view, on a C++ implementation and Intel i7 CPU (2.9GHz).
The model accelerated by our asymmetric solver (channel-only) has 16.7\% error, and by our asymmetric solver (3d) has 14.1\% error. This means that the accelerated model is 4.7\% more accurate than AlexNet, while its speed is nearly the same as AlexNet.

As a common practice \cite{Krizhevsky2012}, we also evaluate the 10-view score of the models. Our accelerated model achieves 12.0\% error, which means only \textbf{0.9\%} increase of error with 4$\times$ speedup (the original one has 11.1\% 10-view error).

\setlength{\tabcolsep}{4pt}
\renewcommand{\arraystretch}{1.2}
\begin{table}
\begin{center}
\begin{small}
\begin{tabular}{|c|c|c|c|c|}
\hline
model & \footnotesize{\tabincell{c}{ speedup \\ solution}} & \footnotesize{\tabincell{c}{top-5 err.\\(1-view)}} & \footnotesize{\tabincell{c}{top-5 err.\\(10-view)}} & \footnotesize{\tabincell{c}{time\\(ms)}} \\
\hline
\emph{AlexNet} \cite{Krizhevsky2012} & - & 18.8 & 16.0 & 273 \\
\hline
\multirow{3}{*}{\tabincell{c}{SPPnet\\\footnotesize{(Overfeat-7)}}}  & \cite{Jaderberg2014}, 4$\times$ & 18.5 & 15.6& 278 \\
                                                      & our asym., 4$\times$ & \textbf{16.7} & \textbf{14.4} & 271 \\
                                                      & our asym. (3d), 4$\times$ & \textbf{14.1} & \textbf{12.0} & 267 \\
\hline
\end{tabular}
\end{small}
\end{center}
\caption{Comparisons of network performance. The top-5 error is absolute values (not the increased number). The running time is per view on a CPU (single thread, with SSE).}
\label{tbl:alexnet}
\end{table}

\section{Conclusion and Future Work}

On the core of our algorithm is the low-rank constraint. While this constraint is designed for speedup in this work, it can be considered as a regularizer on the convolutional filters. We plan to investigate this topic in the future.

{\small
\bibliographystyle{ieee}
\bibliography{cnn_speedup}

\begin{thebibliography}{10}\itemsep=-1pt

\bibitem{Chatfield2014}
K.~Chatfield, K.~Simonyan, A.~Vedaldi, and A.~Zisserman.
\newblock Return of the devil in the details: Delving deep into convolutional
  nets.
\newblock In {\em BMVC}, 2014.

\bibitem{Deng2009}
J.~Deng, W.~Dong, R.~Socher, L.-J. Li, K.~Li, and L.~Fei-Fei.
\newblock Imagenet: A large-scale hierarchical image database.
\newblock In {\em CVPR}, 2009.

\bibitem{Denton2014}
E.~Denton, W.~Zaremba, J.~Bruna, Y.~LeCun, and R.~Fergus.
\newblock Exploiting linear structure within convolutional networks for
  efficient evaluation.
\newblock In {\em NIPS}, 2014.

\bibitem{Girshick2014}
R.~Girshick, J.~Donahue, T.~Darrell, and J.~Malik.
\newblock Rich feature hierarchies for accurate object detection and semantic
  segmentation.
\newblock In {\em CVPR}, 2014.

\bibitem{Golub1996}
G.~H. Golub and C.~F. van Van~Loan.
\newblock Matrix computations.
\newblock 1996.

\bibitem{Gower2004}
J.~C. Gower and G.~B. Dijksterhuis.
\newblock {\em Procrustes problems}, volume~3.
\newblock Oxford University Press Oxford, 2004.

\bibitem{Hariharan2014}
B.~Hariharan, P.~Arbel{\'a}ez, R.~Girshick, and J.~Malik.
\newblock Simultaneous detection and segmentation.
\newblock In {\em ECCV}, pages 297--312, 2014.

\bibitem{He2014}
K.~He, X.~Zhang, S.~Ren, and J.~Sun.
\newblock Spatial pyramid pooling in deep convolutional networks for visual
  recognition.
\newblock {\em arXiv:1406.4729v2}, 2014.

\bibitem{Howard2013}
A.~G. Howard.
\newblock Some improvements on deep convolutional neural network based image
  classification.
\newblock In {\em arXiv:1312.5402}, 2013.

\bibitem{Jaderberg2014}
M.~Jaderberg, A.~Vedaldi, and A.~Zisserman.
\newblock Speeding up convolutional neural networks with low rank expansions.
\newblock In {\em BMVC}, 2014.

\bibitem{Krizhevsky2012}
A.~Krizhevsky, I.~Sutskever, and G.~Hinton.
\newblock Imagenet classification with deep convolutional neural networks.
\newblock In {\em NIPS}, 2012.

\bibitem{LeCun1989}
Y.~LeCun, B.~Boser, J.~S. Denker, D.~Henderson, R.~E. Howard, W.~Hubbard, and
  L.~D. Jackel.
\newblock Backpropagation applied to handwritten zip code recognition.
\newblock {\em Neural computation}, 1989.

\bibitem{Lin2013}
M.~Lin, Q.~Chen, and S.~Yan.
\newblock Network in network.
\newblock In {\em arXiv:1312.4400}, 2013.

\bibitem{Nair2010}
V.~Nair and G.~E. Hinton.
\newblock Rectified linear units improve restricted boltzmann machines.
\newblock In {\em ICML}, pages 807--814, 2010.

\bibitem{Reeves1993}
C.~R. Reeves.
\newblock {\em Modern heuristic techniques for combinatorial problems}.
\newblock John Wiley \& Sons, Inc., 1993.

\bibitem{Russakovsky2014}
O.~Russakovsky, J.~Deng, H.~Su, J.~Krause, S.~Satheesh, S.~Ma, Z.~Huang,
  A.~Karpathy, A.~Khosla, M.~Bernstein, et~al.
\newblock Imagenet large scale visual recognition challenge.
\newblock {\em arXiv:1409.0575}, 2014.

\bibitem{Sermanet2014}
P.~Sermanet, D.~Eigen, X.~Zhang, M.~Mathieu, R.~Fergus, and Y.~LeCun.
\newblock Overfeat: Integrated recognition, localization and detection using
  convolutional networks.
\newblock 2014.

\bibitem{Simonyan2014}
K.~Simonyan and A.~Zisserman.
\newblock Very deep convolutional networks for large-scale image recognition.
\newblock {\em arXiv:1409.1556}, 2014.

\bibitem{Szegedy2014}
C.~Szegedy, W.~Liu, Y.~Jia, P.~Sermanet, S.~Reed, D.~Anguelov, D.~Erhan,
  V.~Vanhoucke, and A.~Rabinovich.
\newblock Going deeper with convolutions.
\newblock {\em arXiv:1409.4842}, 2014.

\bibitem{Takane2007}
Y.~Takane and H.~Hwang.
\newblock Regularized linear and kernel redundancy analysis.
\newblock {\em Computational Statistics \& Data Analysis}, pages 394--405,
  2007.

\bibitem{Takane2006}
Y.~Takane and S.~Jung.
\newblock Generalized constrained redundancy analysis.
\newblock {\em Behaviormetrika}, pages 179--192, 2006.

\bibitem{Vanhoucke2011}
V.~Vanhoucke, A.~Senior, and M.~Z. Mao.
\newblock Improving the speed of neural networks on {CPUs}.
\newblock In {\em Deep Learning and Unsupervised Feature Learning Workshop,
  NIPS 2011}, 2011.

\bibitem{Wang2010}
J.~Wang, J.~Yang, K.~Yu, F.~Lv, T.~Huang, and Y.~Gong.
\newblock Locality-constrained linear coding for image classification.
\newblock In {\em CVPR}, 2010.

\bibitem{Zeiler2014}
M.~D. Zeiler and R.~Fergus.
\newblock Visualizing and understanding convolutional neural networks.
\newblock In {\em ECCV}, 2014.

\end{thebibliography}
}

\end{document}